\documentclass[sigconf]{acmart}

\usepackage{amsmath}
\usepackage{amsfonts}
\usepackage{algorithmic}
\usepackage{graphicx}
\usepackage{textcomp}
\usepackage{xcolor}
\usepackage{filecontents}  
\usepackage{booktabs} 
\usepackage{multirow} 

\usepackage{array}
\usepackage{booktabs} 
\usepackage{float} 
\usepackage{tabularx} 
\usepackage{multirow} 
\usepackage{subcaption} 
\usepackage{pgfplots} 
\pgfplotsset{compat=1.18}
\usepackage{soul} 
\usepackage{caption} 
\usepackage{tikz} 
\usepackage{adjustbox} 

\AtBeginDocument{%
  }

\copyrightyear{2025}
\acmYear{2025}
\setcopyright{acmlicensed}\acmConference[ICMR '25]{Proceedings of the 2025 International Conference on Multimedia Retrieval}{June 30-July 3, 2025}{Chicago, IL, USA}
\acmBooktitle{Proceedings of the 2025 International Conference on Multimedia Retrieval (ICMR '25), June 30-July 3, 2025, Chicago, IL, USA}
\acmDOI{10.1145/3731715.3733327}
\acmISBN{979-8-4007-1877-9/2025/06}
\begin{document}

\title{Enhancing Self-Supervised Fine-Grained Video Object Tracking with Dynamic Memory Prediction}

\author{Zihan Zhou}
\authornote{Both authors contributed equally to this research.}
\orcid{0009-0007-3650-5464}
\affiliation{%
  \department{School of Computer Science and Engineering,}
  \institution{Southeast University,}
  \city{Nanjing}
  \country{China}
}
\email{zhzhou@seu.edu.cn}

\author{Changrui Dai}
\authornotemark[1]
\orcid{0009-0007-9417-9599}
\affiliation{%
  \department{School of Computer Science and Engineering,}
  \institution{Southeast University,}
  \city{Nanjing}
  \country{China}
}
\email{crdai@seu.edu.cn}

\author{Aibo Song}
\orcid{0000-0003-4447-7305}
\affiliation{%
  \department{School of Computer Science and Engineering,}
  \institution{Southeast University,}
  \city{Nanjing}
  \country{China}
}
\affiliation{
  \institution{Key Laboratory of Computer Network and Information Integration (Southeast University), Ministry of Education}
  \city{Nanjing}
  \country{China}
}
\email{absong@seu.edu.cn}

\author{Xiaolin Fang}
\authornote{Corresponding author}
\orcid{0000-0002-0164-2596}
\affiliation{%
  \department{School of Computer Science and Engineering,}
  \institution{Southeast University,}
  \city{Nanjing}
  \country{China}
}
\affiliation{
  \institution{Key Laboratory of Computer Network and Information Integration (Southeast University), Ministry of Education}
  \city{Nanjing}
  \country{China}
}
\email{xiaolin@seu.edu.cn}

\renewcommand{\shortauthors}{Zihan Zhou et al.}

\begin{abstract}
Successful video analysis relies on accurate recognition of pixels across frames, and frame reconstruction methods based on video correspondence learning are popular due to their efficiency. Existing frame reconstruction methods, while efficient, neglect the value of direct involvement of multiple reference frames for reconstruction and decision-making aspects, especially in complex situations such as occlusion or fast movement. In this paper, we introduce a Dynamic Memory Prediction (DMP) framework that innovatively utilizes multiple reference frames to concisely and directly enhance frame reconstruction. Its core component is a Reference Frame Memory Engine that dynamically selects frames based on object pixel features to improve tracking accuracy. In addition, a Bidirectional Target Prediction Network is built to utilize multiple reference frames to improve the robustness of the model. Through experiments, our algorithm outperforms the state-of-the-art self-supervised techniques on two fine-grained video object tracking tasks: object segmentation and keypoint tracking.
\end{abstract}



\maketitle

\begin{figure}[t]
\raggedright
\begin{tikzpicture}
\begin{axis}[
    width=0.99\columnwidth, 
    xlabel={Number of pixel-level annotations (log scale)},
    ylabel={$\mathcal{J} \& \mathcal{F}(Mean)$},
    xmode=log, 
    xmin=1, xmax=10000000, 
    xticklabel style={align=center, text width=1.5cm, anchor=south, yshift=-3.0ex,font=\scriptsize},
    xticklabels={$0$, $10^1$, $10^2$, $10^3$, $10^4$, $10^5$, $10^6$, $10^7$},
    yticklabel style={font=\scriptsize},
    xlabel style={font=\small}, 
    ylabel style={font=\small},
    ymin=63, ymax=87, 
    grid=both,
    grid style={dotted},
    enlargelimits={abs=0.1},
    only marks, 
    mark size=1.8, 
    axis line style={draw=none}    
]

    \addplot[color=blue, mark=*] table {
    x   y
    1   65.5    
    1   66.7    
    1   68.3       
    1   70.3    
    1   71.4    
    1   72.1
    1   74.5
    };
    \addplot[only marks, mark=*, mark size=1.6, fill=red, draw=red] table{
    x   y
    1   76.4   
    };
    \addplot[color=yellow,mark=*] table {
    x   y
    5000000 85.4    
    2000000 82.8    
    3000000 81.7
    };
    \addplot[color=green,mark=*] table {
    x   y
    300000 75.9  
    600000 74.6  
    200000 75.3
    };
    \addplot[color=purple,mark=*] table {
    x   y
    200000 70.0 
    30000 66.7   
    };
    \node[anchor=west, font=\tiny] at  (axis cs:1,   65.5) {MAST\cite{Lai_2020_CVPR}};
    \node[anchor=west, font=\tiny] at  (axis cs:1,   66.7) {VFS\cite{xu2021rethinking}};
    \node[anchor=west, font=\tiny] at  (axis cs:1,   68.3) {CRW\cite{jabri2020space}};
    \node[anchor=west, font=\tiny] at  (axis cs:1,   70.3) {CLTC\cite{jeon2021mining}};
    \node[anchor=west, font=\tiny] at  (axis cs:1,   71.4) {DINO\cite{caron2021emerging}};
    \node[anchor=west, font=\tiny] at  (axis cs:1,   72.1) {LIIR\cite{li2022locality}};
    \node[anchor=west, font=\tiny] at  (axis cs:1,   74.5) {MASK-VOS\cite{li2023unified}};
    \node[anchor=west, font=\tiny] at  (axis cs:1,   76.4) {\textbf{Ours}};
    \node[anchor=east, font=\tiny] at  (axis cs:5000000, 85.4) {STCN\cite{cheng2021rethinking}};
    \node[anchor=east, font=\tiny] at  (axis cs:2000000, 82.8) {EGMN\cite{lu2020video}};
    \node[anchor=east, font=\tiny] at  (axis cs:3000000, 81.7) {STM\cite{oh2019video}};
    \node[anchor=west, font=\tiny] at  (axis cs:300000, 75.9) {Fasttan\cite{huang2020fast}};
    \node[anchor=west, font=\tiny] at  (axis cs:600000, 74.6) 
    {AFB-URR\cite{liang2020video}};
    \node[anchor=east,font=\tiny] at  (axis cs:200000, 75.3) 
    {MHP-VOS\cite{xu2019mhp}};
    \node[anchor=west, font=\tiny] at  (axis cs:200000, 70.0) {AGAME\cite{johnander2019generative}};
    \node[anchor=west, font=\tiny] at  (axis cs:30000, 66.7) {RGMP\cite{oh2018fast}};
\end{axis}
\end{tikzpicture}
\caption{\textbf{Performance comparison} over DAVIS$_{17}$\protect\cite{perazzi2016benchmark}val. The self-supervised method is positioned on the far left of the figure, while the supervised method is located on the right side. Ours surpasses all existing self-supervised methods ($\mathcal{J} \& \mathcal{F}(Mean)$ : 76.4), and is on par with many fully-supervised ones trained with massive annotations.
}
\label{niubi}
\end{figure}
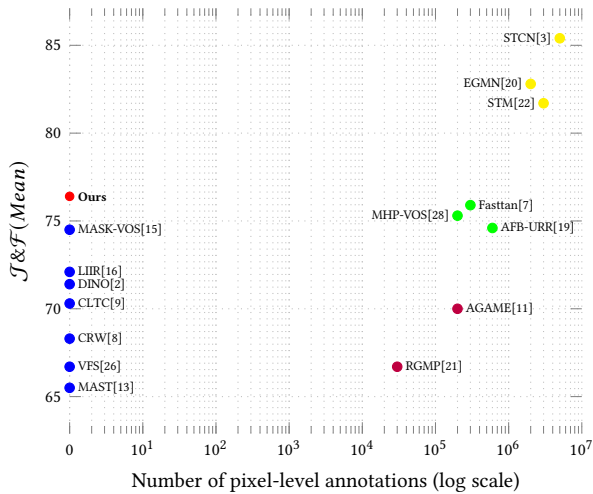

\section{Introduction}
\label{intro}

Unlike supervised video object tracking learning, self-supervised methods alleviate the need for fine pixel-level annotation of video objects and can quickly and cheaply build mature and usable models, which has significant advantages. Due to the inherent temporal and spatial consistency of natural video sequences, utilizing unlabeled natural videos for self-supervised video object tracking has become a feasible solution. Furthermore, rich natural video information enhances the generalization ability of models, potentially surpassing supervised methods in terms of information availability and applicability.

It's worth noting that the Temporal Correspondence Learning method has gained popularity through extensive practical use. However, some researchers have overlooked a fundamental issue: pixel correspondence is not limited to a single or static reference frame similar to the query frame; it spans the entire video sequence.

Admittedly, some researchers have recognized that the quality of frame reconstruction can be improved by using more reference frames, which is usually achieved through methods such as temporal attention mechanisms. While this approach allows for the use of multiple reference frames, it also poses a number of challenges: loss of pixel information during secondary coding, increased training costs, and additional memory overhead. These issues undermine the practical advantages of the approach. Therefore, we design a framework that allows multiple reference frames to participate directly and efficiently in frame reconstruction. The framework encodes a sequence of video frames uniformly once (encoder only), and then enables the model to heuristically select a sufficient number of reference frames in real-time and dynamically during the reconstruction process.

In order to make up for the neglect of multi-frame reference in previous works and enable the model to make full and appropriate use of reference frames, we propose a new frame reconstruction framework, including: 
\begin{itemize}
    \item we propose \textbf{Reference Frame Memory Engine}, which manages two memory banks: a Short-Term Memory bank comprising reference frames temporally proximate to the current query frame and exhibiting high similarity to it, and a Long-Term Memory bank storing reference frames temporally distant from the query frame but containing crucial initial target pixel information for tracking. Both memories are updated dynamically.
    \item we propose \textbf{Bidirectional Target Prediction Network} to handle both long-term and short-term reference frames separately. This approach through building cluster models to establish region correspondence between different frames. For short-term reference frames, we distinguish pixels of different targets through backward-to-forward matching to achieve more precise affinity calculations. For long-term reference frames, we constrain the range of candidate regions for query pixels through forward-to-backward matching, helping to suppress offset during label propagation.
\end{itemize}

Overall, we propose a new framework for self-supervised video correspondence learning named DMP (Dynamic Memory Prediction). This method introduces rich tracking object pixel information through the Reference Frame Memory Engine, and makes good use of this information through the Bidirectional Target Prediction Network. This allows our framework to show excellent accuracy and robustness, surpassing other methods in the same field. And in certain cases, it shows comparable or adaptable performance to fully supervised methods, as shown in Figure ~\ref{niubi}.

\section{Related Work}
\label{related}

\subsection{Self-Supervised Temporal Correspondence Learning}
\label{Reference}
In the field of video processing, understanding correspondence plays a crucial role in a variety of tasks including video segmentation \cite{Hu_2018_ECCV}, flow estimation \cite{dosovitskiy2015flownet,zhao2022global}, and video object tracking \cite{bertinetto2016fully}. Unlike traditional supervised methods, self-supervised approaches for correspondence learning have capitalized on diverse and abundant unlabeled video datasets \cite{2018Geometry,kim2019self,vondrick2018tracking} to discover latent correspondence patterns, leading to substantial advancements.

The dominant approach of self-supervised temporal correspondence learning methods aims at reconstructing the query frame from neighboring frames using inter-frame pixel similarity. \cite{Lai_2020_CVPR,lai2019self,vondrick2018tracking} introduces a colorization proxy task, aiming to reconstruct a query frame from its adjacent frame while leveraging their temporal correspondence. 

Videos are advantageous due to their inherent temporal information and spatio-temporal consistency. This allows us to use the continuity of pixel information, as pixels in one frame often correspond to pixels in the next. Based on this correlation, studies like \cite{lai2019self} have introduced reconstruction methods where each pixel in the current frame identifies its best match in the previous frame.

This reconstruction method incorporates attention mechanisms. Both the current frame and the previous frame undergo feature encoding and are projected into the pixel embedding space $\Phi:R^{H \times W \times 3} \rightarrow R^{h \times w \times c}$, yielding a triplet for each frame $I_{t}$, denoted as ${{Q_{t}, K_{t}, V_{t}}}$, where $Q_t = K_t = \Phi(I_t;\theta)$. During the training phase, the value $V_t$ corresponds to the original reference frame, while during the inference phase, it corresponds to the instance segmentation mask. 

To reconstruct pixel \( i \) in the current frame \( I_t \), we perform similarity sampling from pixels in previous frames. Matching pixels contribute to reconstructing the pixel \( \hat{i} \) in the reconstructed frame \( \hat{I_t} \). This process involves using multiple previous frames and the current frame to compute affinity matrices and generate reconstruction results, summarized by:
\begin{equation}
    A^{ij}_{t, r} = \frac{\exp \langle Q^{i}_{t}, K^{j}_{r} \rangle}{\sum_{p} \exp \langle Q^{i}_{t}, K^{p}_{r} \rangle},
\label{base1}
\end{equation}
where \( A^{ij}_{t, r} \) denotes the affinity between the \( i \)-th feature of the target frame and the \( j \)-th feature of the \( r \)-th reference frame, with \( \langle Q^{i}_{t}, K^{j}_{r} \rangle \) representing their dot product.
\begin{equation}
    \hat{I}^{i}_{t, r} = \sum_{j} A^{ij}_{t, r} V^{j}_{r},
\label{base2}
\end{equation}
where \( \hat{I}^{i}_{t, r} \) is the reconstructed value at pixel \( i \) using features from the \( r \)-th reference frame, and \( V^{j}_{r} \) is the value associated with the \( j \)-th feature in the \( r \)-th reference frame.

\begin{figure*}[h]
    \centering
    \includegraphics[width=0.99\textwidth]{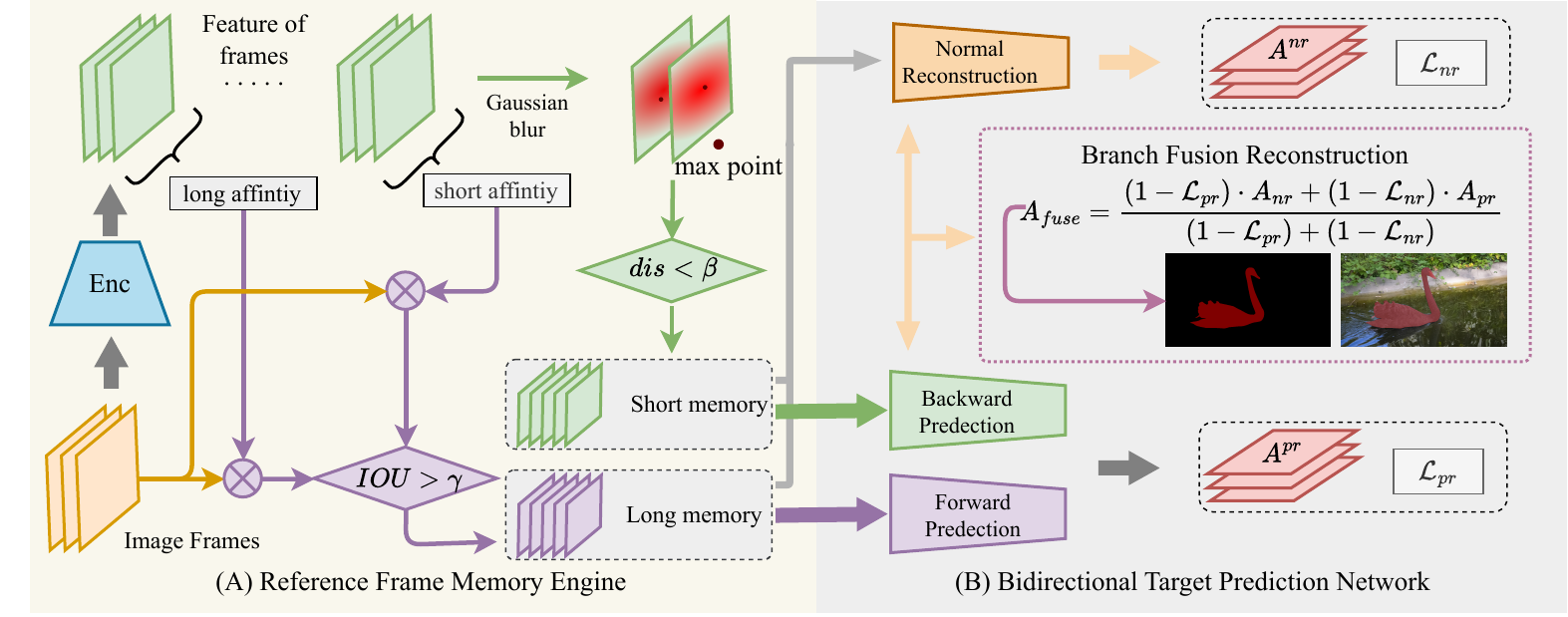}
    \caption{The model architecture of DMP. In our encoder-only model, reference frames are dynamically placed in Long-Term or Short-Term Memory. Frames in different memory banks will perform forward or backward target prediction with the target frame separately to obtain the best tracking results.}
    \label{fig:baseline}
\end{figure*}

\section{Method}
\label{sec:method}

We introduce an innovative multi-frame referencing approach for frame reconstruction , dynamically selecting multiple reference frames to engage in the process (As shown in \textbf{Reference Frame Memory Engine}). This enriches video object information, enhancing tracking performance. Furthermore, our method capitalizes on multi-frame reference potential by not only involving reference frames in reconstruction but also evaluating the reconstruction outcomes (As shown in \textbf{Bidirectional Target Prediction Network}), thus bolstering tracking result robustness. The model architecture can be found in Fig.~\ref{fig:baseline}.

\subsection{Reference Frame Memory Engine}
As shown in \ref{Reference}, during the model's inference about the object mask in the query frame $I_t$, it is necessary to sample similarity information from the reference frame for reconstruction. Unlike previous network-based memorization mechanisms, we propose a heuristic method to reconstruct multiple reference frames, aiming to create a simple yet efficient reference frame memory engine without unnecessary processing.

When introducing multi-frame referencing, we found that the contribution of previous frames to the reconstruction varies significantly. The reason for this uneven contribution is the different time intervals relative to the query frame, and each frame plays a different specific role in video object tracking. Frames temporally closer to the query frame exhibit higher similarity and provide rich pixel correspondences. In contrast, initial video frames offer essential pixel information about the object, which aids in accurate localization and addresses challenges such as occlusion and rapid movements. Together, these complementary contributions enable robust tracking across varying scenarios. So as shown in Fig.~\ref{fig:baseline}(A), we focus on the entire video sequence and establish a Reference Frame Memory Engine based on the standard of whether the reference frame contains pixel information belonging to the tracking object. This engine maintains reference frames with different contributions through two modules.

The reference frame memory engine uses two memory banks: short-term memory and long-term memory. Short-term memory stores frames close to the query frame, which, due to their temporal proximity, provide detailed pixel information, thereby improving reconstruction accuracy. However, as tracking continues, errors in pixel reconstruction may accumulate. Long-term memory helps offset this by storing frames far from the query frame, capturing raw pixel information to maintain tracking stability and mitigate error accumulation.

\subsubsection{Short-Term Memory}
The inherent spatio-temporal coherence of video is manifest in the high similarity between frames that are in close temporal proximity. Consequently, the principles that dictate Short-Term Memory can be inferred from this characteristic.

Video frames all pass through the feature extractor $\Phi$. $\Phi$ extracts the pixel features of the object. Input the video frame to the feature extractor $\Phi$ to obtain the feature map. After performing two-dimensional Gaussian blur on the feature map, the maximum value point reveals exactly the location of the target. The position $p_{t-1}, p_{t-2}, p_{t-3}, \cdots, p_{t-n}$, obtained by the above method in multiple previous frames $I_{t-1}, I_{t-2}, I_{t-3}, \cdots, I_{t-n}$. If the distance between the positions $p_t$ obtained by the current frame $I_{t}$ is within a certain threshold $\beta$ (Adaptive hyperparameters, with initial values set to 0.15), then these previous frames can be considered qualified to become the reference frames of the current frame and added to the Short-Term Memory. This process can be summarized as follows:
\begin{equation}
    f\left(x,y\right)=\frac{1}{2\pi\sigma_x\sigma_y}\exp{\left(-\frac{\left(x-\mu_x\right)^2}{2\sigma_x^2}-\frac{\left(y-\mu_y\right)^2}{2\sigma_y^2}\right)},
\end{equation}
\begin{equation}
    p = \underset{(x,y)}{\text{argmax}} f(x,y),
\end{equation}
\begin{equation}
    dis\left(p_{t},p_{r}\right)<\beta ,  r \in \{0,1,...,t-1\}.
\end{equation}
Here, $f$ denotes the two-dimensional Gaussian blur applied to frame $I_t$, and $p_{t}$ identifies the peak location in the resulting feature map, representing the tracked object's position.. $p_{r}$ represents the position of the maximum value within $I_{r}$ after undergoing the two-dimensional Gaussian blur. $dis(p_{t},p_{r})$ represents the Euclidean distance between $p_{t}$ and $p_{r}$, any $I_{r}$ with $dis(p_{t},p_{r})<\beta$ (In this work, $\beta$ is set to 0.15) will be put into Short-Term Memory. 

\subsubsection{Long-Term Memory}
The introduction of Long-Term Memory frames ensure the model "doesn't lose sight of its original objective." For example, the initial frame ($I_0$) provides the foundational pixel information for tracking. By integrating Long-Term Memory frames into the reconstruction process, the system can reliably pinpoint the target, compensating for the limitations of short-term correspondence learning. This approach addresses challenges from rapid target movements and error accumulation, enhancing the stability and reliability of reconstruction outcomes.

However, Long-Term Memory frames often differ significantly from the current frame, containing both relevant and extraneous information. Incorporating them without careful selection may introduce noise or irrelevant information, potentially degrading the reconstruction performance. Therefore, this article uses a direct method: reconstruct a frame at least a long enough frames away from the current frame to obtain the reconstruction result $\hat{I}_{long}$. Then, reconstruct the Short-Term Memory frame and the current frame to get $\hat{I}_{short}$. Compare $\hat{I}_{short}$ with $\hat{I}_{long}$ using the intersection-over-union (IoU) ratio. Based on this comparison, determine which frame to include in Long-Term Memory. This mechanism allows Long-Term Memory frames to complement Short-Term Memory by providing stable references, thereby enhancing adaptability to rapid motion and significant appearance variations in the current frame. The formula is as follows:
\begin{equation}
    IoU = \frac{\hat{I}_{long}\cap \hat{I}_{short}}{\hat{I}_{long}\cup \hat{I}_{short}}.
\end{equation}
Here, $\hat{I}$ represents the result of reconstructing the current frame using Formulas \eqref{base1}, \eqref{base2}. The object mask in the reconstructed frame will participate in the IoU calculation. We establish a threshold $\gamma$: if the $IoU > \gamma$ (Adaptive hyperparameters, with initial values set to 0.85), the distant frame is deemed suitable for inclusion in the Long-Term Memory.

It is worth mentioning that we need to limit these two memory buffers to prevent them from expanding during the reconstruction process, which can cause heavy computational and storage burdens. Thus, dynamic pruning is necessary. A straightforward approach is to set a memory length limit and use a first-in-first-out (FIFO) policy when the limit is reached. In addition, we have developed a pruning method that evaluates the affinity between the current frame and the memory frame based on the feature data distribution. During the frame reconstruction process with the memory frames, we calculate the Fréchet Inception Distance (FID) between the current frame and each frame in the memory. The FID serves as a score for the frames in memory, and the frame with the highest FID score is then removed. The formula is as follows:
\begin{equation}
\text{FID} = \left\| \mu_t - \mu_r \right\|^2_2 + \text{tr}\left( \Sigma_t + \Sigma_r - 2\left(\Sigma_t\Sigma_r\right)^{\frac{1}{2}} \right),
\end{equation}
$\mu_t$ and $\Sigma_t$ represent the mean and covariance matrix of the pixels in the current frame, respectively, while $\mu_r$ and $\Sigma_r$ represent the mean and covariance matrix of the pixels in the reference frame stored in memory, respectively. At this point, Reference Frame Memory Engine incorporates a dynamic update mechanism, enabling efficient frame addition and removal while maintaining reconstruction performance and computational efficiency.

\begin{figure}[h]
    \centering
    \includegraphics[width=0.99\linewidth]{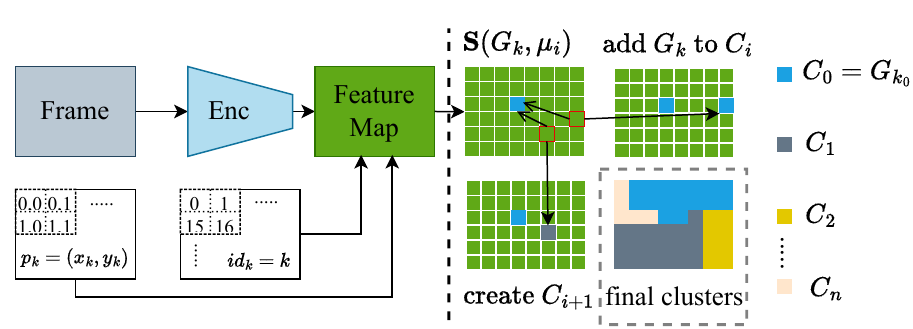}
    \caption{\textbf{Our Frame Region Clustering Model.} Blocks of the same color indicate that these regions belong to the same cluster.}
    \label{cluster model fig}
\end{figure}

\subsection{Bidirectional Target Prediction Network}
\subsubsection{Frame Region Clustering} \label{3.3.1 Frame Region Clustering}
To leverage the inherent spatial consistency of natural video frames, we propose a cross-frame spatial correspondence algorithm that establishes coarse-grained correspondences between frames. Given a reference frame \( I_{r} \) and a target frame \( I_{t} \), our algorithm divides both frames into grids, clusters these grids into groups of similar features, and aligns spatially similar regions across frames. The algorithm consists of two main steps:

\textbf{Step 1: Target Frame Clustering.}  
Firstly, we extract the target frame's feature map \( F_{t} = \Phi(I_{t}) \in \mathbb{R}^{h \times w \times c} \) using a feature extractor \( \Phi \) , then partition it into \( n \) non-overlapping grids  \( f_k \in \mathbb{R}^{s \times s \times c} \),  satisfying:
\begin{equation}
\bigcup_{k=1}^{n} f_k = F_{t}, \quad f_i \cap f_j = \emptyset \, \text{for } i \neq j.
\end{equation}

Each grid is augmented with its center position \( p_k = (x_k, y_k) \) and sequence number \( id_k = k \), forming \( G_k = (f_k, p_k, id_k) \). These grids are clustered into groups \( C = \{C_1, \ldots, C_m\} \), where each cluster \( C_i \) is represented by the average feature vector \( \mu_{f_i} \) and position \( \mu_{p_i} \):
\[
C_i = \{G_{k_1}, G_{k_2}, \ldots, G_{k_{n_i}}\},
\]
\begin{equation}
\mu_{f_i} = \frac{1}{n_i} \sum_{j=1}^{n_i} f_{k_j}, \quad \mu_{p_i} = \frac{1}{n_i} \sum_{j=1}^{n_i} p_{k_j},
\end{equation}
where \( n_i \) is the number of grids in cluster \( C_i \).

The clustering algorithm begins by selecting an initial grid \( G_{k_0} \) as the first cluster center. For each remaining grid, its similarity to existing clusters is computed as:
\begin{equation}
S(G_k, C_i) = \frac{1}{1 + \lambda \|f_k - \mu_{f_i}\|_2 + (1-\lambda) \|p_k - \mu_{p_i}\|_2},
\end{equation}
where \( \lambda \in [0, 1] \) balances feature and positional similarity. The maximum similarity \( S_{max} \), mean similarity \( \bar{S} \), and standard deviation \( S_{\sigma} \) are calculated. If \( S_{max} > \bar{S} + 2S_{\sigma} \), the grid is assigned to the most similar cluster; otherwise, a new cluster is created (see Fig.~\ref{cluster model fig}, right).

\textbf{Step 2: Reference Frame Clustering.}  
The reference frame is processed similarly. Its grids are clustered using the same method but without creating new clusters. Grids with \( S_{max} \leq \bar{S} + 2S_{\sigma} \) are discarded, as they do not correspond to the target frame.

\begin{table*}[t] 
\renewcommand\arraystretch{1.4} 
\centering
\begin{adjustbox}{width=\textwidth}
\begin{tabular}
{>{\centering\arraybackslash}m{3.0cm}
    >{\centering\arraybackslash}m{1.6cm}
    >{\centering\arraybackslash}m{3.0cm}
    >{\centering\arraybackslash}m{1.6cm}
    >{\centering\arraybackslash}m{1.6cm}
    >{\centering\arraybackslash}m{1.6cm}
    >{\centering\arraybackslash}m{1.6cm}
    >{\centering\arraybackslash}m{1.6cm}
    }
\toprule 
Method & Backbone & Dataset & $\mathcal{J}$\&$\mathcal{F}_{m}\uparrow$ & $\mathcal{J}(Mean)\uparrow$ & $\mathcal{J}$(Recall)$\uparrow$ & $\mathcal{F}(Mean)\uparrow$ & $\mathcal{F}$(Recall)$\uparrow$ \\

\midrule 
MAST\cite{Lai_2020_CVPR}                  &ResNet-18   &YouTube-VOS         & 65.5 & 63.3 & 73.2 & 76.6 & 77.7 \\ 
CRW\cite{jabri2020space}                  &ResNet-18   &Kinetics          & 68.3 &65.5 &78.6 &71.0 &82.9  \\
JSTG\cite{zhao2021modelling}              &ResNet-18    &Kinetics             &68.7 &65.8 &77.7 &71.6 &84.3\\
CLTC\cite{jeon2021mining}                 &ResNet-18    &YouTube-VOS       &70.3 &67.9 &78.2 &72.6 &83.7 \\
LIIR\cite{li2022locality}                 &ResNet-18    &YouTube-VOS      &72.1 &69.7 &81.4 &74.5 &85.9\\
FGVC\cite{li2023learning}                 &ResNet-18    &YouTube-VOS       &72.4 &70.5 &- &74.4 &-\\
SpaT\cite{li2023spatial}                  &ResNet-18    &ImageNet+YTV   &73.6 &70.7 &- &76.4 &-\\
MASk-VOS\cite{li2023unified}              &ResNet-18    &YouTube-VOS       &74.5 &71.6   &82.9   &77.4   &86.9\\
\hline
\textbf{DMP}                              &ResNet-18    &YouTube-VOS        &\textbf{76.4} &\textbf{74.6} &\textbf{87.2} &\textbf{78.2} &\textbf{88.5}\\
\bottomrule 
\end{tabular}
\end{adjustbox}
\caption{\textbf{Quantitative results for video object segmentation} on DAVIS$_{17}$\protect\cite{perazzi2016benchmark}val.}
\label{Quantitative results for DAVIS}
\end{table*}  

\subsubsection{Forward Target Prediction via Clustered Feature Labels}\label{3.3.2 Forward Target Prediction}

Using the clusters formed in Section~\ref{3.3.1 Frame Region Clustering}, we predict the target frame features from the reference frame. Each grid in the target frame is assigned an initial label: $\ell_k = f_k - \mu_{f_{k}}$, where \( f_k \) is the grid feature, and \( \mu_{f_{k}} \) is the feature center of the corresponding cluster. The labels are optimized by minimizing:
\begin{equation}
    J = \sum_{i=1}^{m} \sum_{f_k \in C_i} \|\ell_{k} - \mu_{f_{k}}\|_2 - \zeta \sum_{k_1 \neq k_2} \|\mu_{f_{k_1}} - \mu_{f_{k_2}}\|_2,
\end{equation}
where \( \zeta \) controls the weight of the inter-cluster regularization term. This objective ensures that features within a cluster remain consistent while features between clusters remain distinct.

The optimized labels are integrated back into the original feature map:
\begin{equation}
    F_{label} = \bigcup_{k=1}^{n} (f_k + \ell_{k}),
\end{equation}
producing an enhanced feature map used for further reconstruction.

This method ensures similarity within clusters and maximizes dissimilarity between different clusters. The enhanced feature map \( F_{label} \) is then used in the affinity matrix calculation (Formula \ref{base1}) to obtain an enhanced affinity matrix.

\begin{figure}[h]
    \centering
    \includegraphics[width=0.99\linewidth]{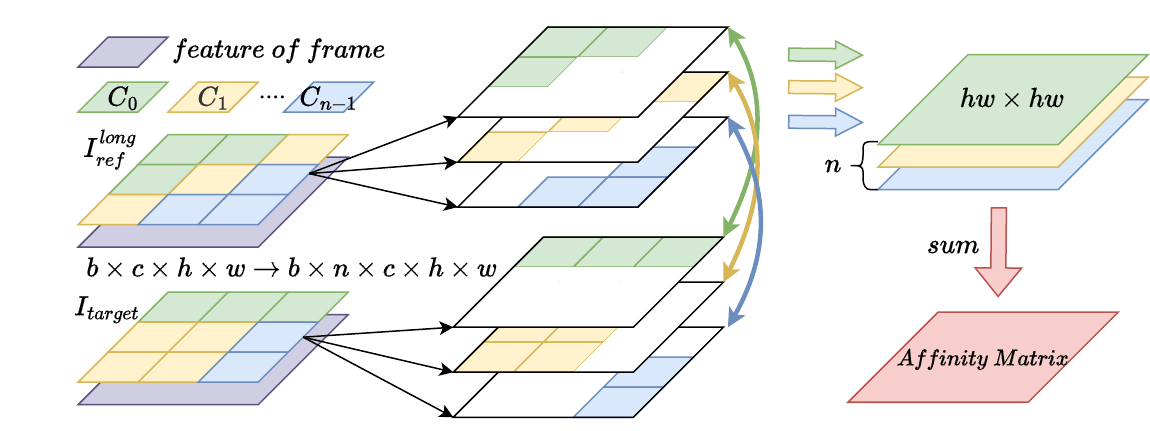}
    \caption{\textbf{Backward Target Prediction.}The candidate regions of pixels in the target frame are restricted to the same color blocks in the reference frames.}
    \label{Backward Target Prediction}  
\end{figure}

\subsubsection{Backward Target Prediction with Restricted Affinity Calculations}\label{3.3.3 Backward Target Prediction}

Backward prediction ensures robust spatial correspondence. Features are split into tensors based on cluster assignments:
\[
F \in \mathbb{R}^{h \times w \times c} \rightarrow F \in \mathbb{R}^{n \times h \times w \times c},
\]
where \( n \) is the number of clusters. Any empty regions within these tensors are padded with zeros  (see Fig.~\ref{Backward Target Prediction}, left). Affinity matrices \( A_i \) are computed for each cluster and merged: $A = \sum_{i=1}^{n} w_i A_i,$ where \( w_i \) is a weight based on cluster similarity. This approach confines affinity calculations to relevant regions, improving accuracy and efficiency.

\subsubsection{Branch Fusion Reconstruction}
We introduce a dual-branch parallel strategy, consisting of a standard reconstruction branch and a Bidirectional Target Prediction Network branch. This approach aims to accelerate the learning of the Bidirectional Target Prediction Network and correct its potential errors using the normal reconstruction branch.

For all reference frames, they are processed in both branches simultaneously, generating affinity matrices \( A_{nr} \) and \( A_{pr} \). These matrices produce the reconstruction results and corresponding losses \( \mathcal{L}_{nr} \) and \( \mathcal{L}_{pr} \). We then combine these losses to form a new affinity matrix:

\begin{equation}
A_{fuse} = \frac{(1-\mathcal{L}_{pr}) \cdot A_{nr} + (1-\mathcal{L}_{nr}) \cdot A_{pr}}{(1-\mathcal{L}_{pr}) + (1-\mathcal{L}_{nr})}
\end{equation}

The final reconstruction result is derived from \( A_{fuse} \), and only the corresponding loss is backpropagated during training (see Fig. \ref{fig:baseline}, right). This dual-branch strategy effectively leverages the complementary strengths of both branches, improving overall model efficiency and accuracy.

To analyze the sensitivity of our model to hyperparameters $\beta$ and $\gamma$, we conduct an ablation study by varying these parameters and evaluating the $\mathcal{J}$\&$\mathcal{F}(Mean)$ Score. To ensure the robustness of the hyperparameter settings, we conducted extensive experiments on a large number of videos with a balanced distribution of categories. The results are shown in Figure~\ref{fig:beta_gamma_sensitivity}, where the x axis (bottom) represents the values $\beta$ and the x axis (top) represents the values $\gamma$.

\begin{figure*}[h]
    \centering
    \begin{subfigure}{\textwidth}
        \centering
        \includegraphics[width=\textwidth]{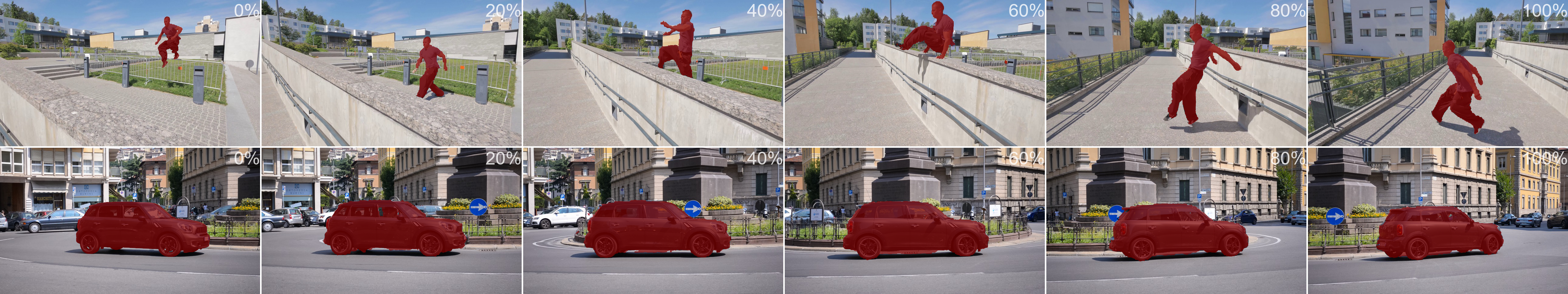}
        \caption{Visualization results on DAVIS$_{17}$\protect\cite{perazzi2016benchmark}val.}
        \label{davisarray}
    \end{subfigure}    
    
    \begin{subfigure}{\textwidth}
        \centering
        \includegraphics[width=\textwidth]{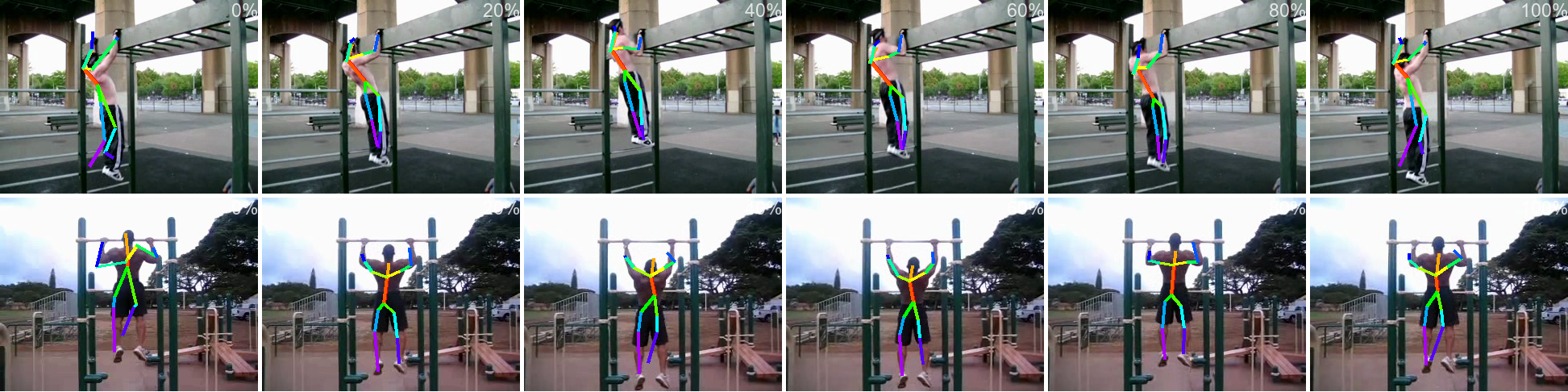}
        \caption{Visualization results for keypoint tracking on JHMDB\protect\cite{jhuang2013towards}val.}
        \label{jhmdbarray}
    \end{subfigure}
    \caption{\textbf{The visualization results of different tasks} on DAVIS$_{17}$\protect\cite{perazzi2016benchmark}val (a) and JHMDB\protect\cite{jhuang2013towards}val (b).}
    \label{visible benchmark}
\end{figure*}

\section{Experiment}
\label{sec:exp}
\subsection{Implementation Details}
To ensure a fair comparison, we used ResNet18 as the feature encoder in all experiments. The model was trained fully self-supervised on two NVIDIA RTX3090s. Self-supervision is reflected in the fact that we project the image into Lab colour space and extract the a-channels from it as pixel labels, thus calculating the label reconstruction loss as in \cite{Lai_2020_CVPR}. No external information is used except the original video sequence (YouTube-VOS\cite{xu2018youtube}).

\begin{table}[t]  
\renewcommand\arraystretch{1.4} 
\centering
\begin{adjustbox}{width=\linewidth}
\begin{tabular}
{>{\centering\arraybackslash}m{3.0cm}
    >{\centering\arraybackslash}m{2.0cm}
    >{\centering\arraybackslash}m{1.0cm}
    >{\centering\arraybackslash}m{2.0cm}
    }
\toprule 
Method &Supervised & Dataset & $PCK@0.1\uparrow$  \\

\midrule 
JSTG\cite{zhao2021modelling}  &    &K &61.4 \\
CLTC\cite{jeon2021mining} &    &Y  &60.5 \\
LIIR\cite{li2022locality} &    &Y  &60.7 \\
SPAT\cite{li2023spatial}  &    &I+Y &63.1 \\
\hline
\textbf{DMP}  &     &Y          &\textbf{64.9} \\
\hline
Thin-Slicing Net\cite{song2017thin} &$\surd$  &- &68.7 \\
\bottomrule 
\end{tabular}
\end{adjustbox}
\caption{\textbf{Quantitative results for human part propagation and pose keypoint tracking}. We show results of self-supervised methods and some supervised methods for comparison. \textcolor{gray}{\footnotesize Note: K: Kinetics, Y: Youtube-VOS, I: ImageNet.}}
\label{Quantitative results for JHMDB}
\end{table}  

\subsection{Comparisons with State-of-the-Art Methods}
\subsubsection{Video Object Segmentation}
In Table \ref{Quantitative results for DAVIS}, we compare our model with previous approaches on the DAVIS$_{17}$\cite{perazzi2016benchmark}val. We assess methods for video object segmentation using three evaluation metrics: the mean region similarity $\mathcal{J}(Mean)$, the mean contour accuracy $\mathcal{F}(Mean)$, and their average $\mathcal{J}$\&$\mathcal{F}_{m}$. It can be observed that our model outperforms all current self-supervised video object segmentation methods in the benchmark test. It surpasses current best-performing self-supervised method, $i.e.$, MASK-VOS\cite{li2023unified} in terms of mean $\mathcal{J}$\&$\mathcal{F}$ (\textbf{76.4} vs. 74.5). Furthermore, our model exhibits a stronger capability to distinguish between the target and the background, resulting in improved tracking performance. 

\subsubsection{Pose Keypoint Tracking}
Table \ref{Quantitative results for JHMDB} presents the outcomes obtained by our model (DMP) on the JHMDB dataset \cite{jhuang2013towards}, showcasing the quantitative results of human part propagation and pose keypoint tracking. Notably, our model surpasses the majority of prior self-supervised video correspondence learning models, achieving state-of-the-art results particularly on the $PCK@0.1$ metric. Moreover, our work demonstrates significant enhancements in performance compared to certain partially supervised video correspondence learning models.

\begin{table}[h] 
\renewcommand\arraystretch{1.2} 

\begin{subtable} {\columnwidth}
  \centering
\begin{adjustbox}{width=\linewidth}
  \begin{tabular}{
    >{\centering\arraybackslash}m{3.5cm}
    >{\centering\arraybackslash}m{3.5cm}
    >{\centering\arraybackslash}m{1cm}
    >{\centering\arraybackslash}m{1cm}}
    \toprule
    \multicolumn{1}{c}{Short-Term} & \multicolumn{1}{c}{Long-Term}  & \multicolumn{2}{c}{DAVIS} \\
    \cline{3-4} 
    \multicolumn{1}{c}{Memory}  & \multicolumn{1}{c}{Memory}  & \multicolumn{1}{c}{ $\mathcal{J}\uparrow$} & \multicolumn{1}{c}{$\mathcal{F}\uparrow$} \\
    \hline
    \hline
    -       &   -         & 68.0     & 72.1 \\
    \hline
    $\surd$ &  -          & 71.7     & 73.5 \\
    -       & $\surd$     & 66.4     & 71.0 \\
    $\surd$ & $\surd$     & \textbf{72.3}     & \textbf{75.2} \\
    \bottomrule
  \end{tabular}
\end{adjustbox}
  \subcaption{Reference Frame Memory Engine}
  \label{subtab:reference-frame-memory-engine}
\end{subtable}

\begin{subtable} {\columnwidth}
  \centering
\begin{adjustbox}{width=\linewidth}
  \begin{tabular}{
    >{\centering\arraybackslash}m{3.5cm}
    >{\centering\arraybackslash}m{3.5cm}
    >{\centering\arraybackslash}m{1cm}
    >{\centering\arraybackslash}m{1cm}}
    \toprule
    \multicolumn{1}{c}{Forward Target} & \multicolumn{1}{c}{Backward Target}  & \multicolumn{2}{c}{DAVIS} \\
    \cline{3-4} 
    \multicolumn{1}{c}{Prediction}    & \multicolumn{1}{c}{Prediction} &
    \multicolumn{1}{c}{ $\mathcal{J}\uparrow$} & \multicolumn{1}{c}{$\mathcal{F}\uparrow$} \\
    \hline
    \hline
    -     &   -         & 68.0     & 72.1 \\
    \hline
    $\surd$ &   -         & 71.7     & 74.4 \\
    -       & $\surd$     & 71.9     & 74.7 \\
    $\surd$ & $\surd$     & \textbf{73.1}     & \textbf{75.3} \\
    \bottomrule
  \end{tabular}
\end{adjustbox}
  \subcaption{Bidirectional Target Prediction Network}
  \label{subtab:bidirectional-target-prediction-network}
\end{subtable}

\begin{subtable} {\columnwidth}
  \centering
\begin{adjustbox}{width=\linewidth}
  \begin{tabular}{
    >{\centering\arraybackslash}m{3.5cm}
    >{\centering\arraybackslash}m{3.5cm}
    >{\centering\arraybackslash}m{1cm}
    >{\centering\arraybackslash}m{1cm}}
    \toprule
    \multicolumn{1}{c}{Reference Frame}  &
    \multicolumn{1}{c}{Bidirectional Target} &
    \multicolumn{2}{c}{DAVIS} \\
    \cline{3-4} 
    \multicolumn{1}{c}{Memory Engine}      & 
    \multicolumn{1}{c}{Prediction Network}  &
    \multicolumn{1}{c}{ $\mathcal{J}\uparrow$} & 
    \multicolumn{1}{c}{$\mathcal{F}\uparrow$}  \\
    \hline
    \hline
    -        &   -       & 68.0     & 72.1\\
    \hline
    $\surd$ &   -       & 72.3    & 75.2\\
    -        & $\surd$   & 73.1     & 75.3\\
    $\surd$ & $\surd$   & \textbf{74.6 }    & \textbf{78.2}\\
    \bottomrule
  \end{tabular}
\end{adjustbox}
  \subcaption{Dynamic Memory Prediction}
  \label{subtab:detailed-analysis}
\end{subtable}
\caption{A set of ablative studies on DAVIS $_{17}$\protect\cite{perazzi2016benchmark}val.}
\label{ablative studies}
\end{table}

\subsection{Ablation Experiment} \label{section 4.2 Diagnostic Experiment}
We first examine the efficacy of  Reference Frame Memory Engine and Bidirectional Target Prediction Network. The results are summarized in Table \ref{ablative studies}(c). In lines 2-3, we demonstrate the individual impact of using specific modules on the overall model performance. In line 4, we combine all parts  and achieve the best performance. This indicates that these modules can complement each other and work synergistically, showcasing the effectiveness of our overall design.

The Reference Frame Memory Engine demonstrates that Short-Term Memory, rich in target pixel information, significantly improves reconstruction accuracy. Long-Term Memory, containing raw pixel information, complements Short-Term Memory and enhances overall results when combined. For the Bidirectional Target Prediction Network, using static frames for reconstruction, both Forward and Backward Target Predictions enhance the model's understanding of target motion. Introducing either Forward or Backward Target Prediction improves performance, and combining both further enhances the model's effectiveness.

\subsection{Hyperparameters sensitivity experiment} \label{section 4.4 Hyperparameters sensitivity experiment}
\begin{figure}[h]
    \centering
    \begin{tikzpicture}
        \begin{axis}[
            width=0.9\columnwidth,
            height=4cm,
            every axis label/.append style={font=\small},
            every tick label/.append style={font=\footnotesize},
            xlabel={$\beta$ (Bottom Axis) and $\gamma$ (Top Axis) Value },
            ylabel={$\mathcal{J} \& \mathcal{F}(Mean)$ },
            xmin=0, xmax=1,
            ymin=72, ymax=78,
            xtick={0,0.25,0.5,0.75,1},
            xticklabels={0.05, 0.10, 0.15, 0.20, 0.25}, 
            ytick={72,73,74,75,76,77,78},
            legend pos=south east,
            grid=both,
            grid style={dotted},
            axis y line=left,
            axis x line=bottom,
            extra x ticks={0,0.25,0.5,0.75,1}, 
            extra x tick labels={0.75, 0.80, 0.85, 0.90, 0.95}, 
            extra x tick style={ticklabel pos=top},
            legend style={fill=none, draw=none, at={(1,1)}, anchor=north east}
        ]
        
        \addplot[smooth,mark=o,blue] plot coordinates {
            (0, 74.2)
            (0.25, 75.3)
            (0.5, 76.4)
            (0.75, 75.1)
            (1, 73.8)
        };
        \addlegendentry{\scriptsize\textcolor{black}{$\beta$ Sensitivity}}

        \addplot[smooth,mark=square,red] plot coordinates {
            (0, 73.5)
            (0.25, 74.8)
            (0.5, 76.4)
            (0.75, 75.0)
            (1, 73.2)
        };
        \addlegendentry{\scriptsize\textcolor{black}{$\gamma$ Sensitivity}}

        \end{axis}
    \end{tikzpicture}
    \caption{Impact of hyperparameters $\beta$ and $\gamma$ on $\mathcal{J} \& \mathcal{F}(Mean)$ . The bottom x-axis represents $\beta$, and the top x-axis represents $\gamma$, both evenly spaced.}
    \label{fig:beta_gamma_sensitivity}
\end{figure}
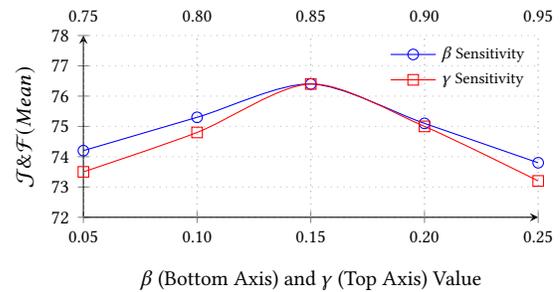

To systematically investigate the sensitivity of our model to hyperparameters $\beta$ and $\gamma$, we conducted an ablation study by varying these parameters and evaluating their influence on the mean $\mathcal{J} \& \mathcal{F}$ metrics on the DAVIS$_{17}$ validation set~\cite{perazzi2016benchmark}. To ensure the robustness and reliability of our analysis, we performed extensive experiments across a large set of videos with a balanced distribution of object categories. The experimental results are illustrated in Figure~\ref{fig:beta_gamma_sensitivity}, where the bottom x-axis denotes the values of $\beta$, and the top x-axis represents the corresponding values of $\gamma$.

\section{Conclusion}
In summary, we propose a multi-reference reconstruction framework based on a Dynamic Memory Prediction method, which we call DMP. This method provides an efficient and concise way to effectively improve the accuracy and robustness of self-supervised fine-grained video object tracking. \textbf{Remarkably}, trained without semantic annotations, our model outperforms previous self-supervised methods on both the video object segmentation benchmark DAVIS$_{17}$\protect\cite{perazzi2016benchmark}val ($\mathcal{J}$\&$\mathcal{F}(Mean)$ is up 2.6\%) and the pose keypoint tracking benchmark JHMDB\protect\cite{jhuang2013towards}val ($PCK@0.1$ is up 2.8\%). This advancement significantly reduces the performance gap with supervised methods, highlighting the thriving potential of enhancing video analysis without relying on semantic annotations. We also believe that our method is decoder-free and directly processes pixel features to obtain results. It has the potential to be plug-and-play in a wider range of video processing fields and is worthy of further exploration.

\section{Acknowledgement}
This research was supported in part by National Natural Science Foundation of China under grant No. 62061146001, 62372102, 62232004, 61972083, 62202096, 62072103, Jiangsu Provincial Key Research and Development Program under grant No. BE2022680, Fundamental Research Funds for the Central Universities (2242024k30022), Jiangsu Provincial Key Laboratory of Network and Information Security under grant No. BM2003201, and Collaborative Innovation Center of Novel Software Technology and Industrialization.

\bibliographystyle{ACM-Reference-Format}
\bibliography{sample-base}

\end{document}